\documentclass[letterpaper, 10 pt, conference]{ieeeconf}  
\usepackage{graphicx}
\usepackage{color}
\usepackage{booktabs}
\usepackage{authblk}
\usepackage{stfloats}
\usepackage{soul}
\usepackage{url}
\usepackage{amssymb}
\usepackage{array}
\usepackage[hidelinks]{hyperref}
\usepackage[utf8]{inputenc}
\usepackage[small]{caption}
\usepackage{graphicx}
\usepackage{amsmath}
\usepackage{amssymb}
\usepackage{algorithm}
\usepackage{algorithmic}
\usepackage[switch]{lineno}
\usepackage{multirow}
\usepackage{xspace}
\usepackage{adjustbox}
\usepackage[bottom=2.0cm, top=2.0cm, left=1.91cm, right=1.91cm]{geometry}
\usepackage{xspace}

\IEEEoverridecommandlockouts                              

\title{\LARGE \bf RTFDNet: Fusion-Decoupling for Robust RGB-T Segmentation}

\author{Kunyu Tan, Mingjian Liang$^{\dagger}$%
\thanks{Kunyu Tan and Mingjian Liang are independent researchers.\\
$^{\dagger}$ Corresponding author: \texttt{2443434059@qq.com}}}

\newcommand{\method}{RTFDNet\xspace}
\begin{document}
\maketitle
\section{Abstract}
\label{sec:abs}

\begin{abstract}

RGB-Thermal (RGB-T) semantic segmentation is essential for robotic systems operating in low-light or dark environments. However, traditional approaches often overemphasize modality balance, resulting in limited robustness and severe performance degradation when sensor signals are partially missing. Recent advances such as cross-modal knowledge distillation and modality-adaptive fine-tuning attempt to enhance cross-modal interaction, but they typically decouple modality fusion and modality adaptation, requiring multi-stage training with frozen models or teacher–student frameworks.
We present \method, a three-branch encoder–decoder that unifies fusion and decoupling for robust RGB-T segmentation. Synergistic Feature Fusion (SFF) performs channel-wise gated exchange and lightweight spatial attention to inject complementary cues. Cross-Modal Decouple Regularization (CMDR) isolates modality-specific components from the fused representation and supervises unimodal decoders via stop-gradient targets. Region Decouple Regularization (RDR) enforces class-selective prediction consistency in confident regions while blocking gradients to the fusion branch. This feedback loop strengthens unimodal paths without degrading the fused stream, enabling efficient standalone inference at test time.
Extensive experiments demonstrate the effectiveness of~\method~, showing consistent performance across varying modality conditions. Our implementation will be released to facilitate further research. Our source code are publicly available at \url{https://github.com/curapima/RTFDNet}.
\end{abstract}
    
\section{Introduction}
\label{sec:intro}
Multimodal perception has become instrumental in Embodied-AI~\cite{li2025bridgevlainputoutputalignmentefficient, yuan2025depthvlaenhancingvisionlanguageactionmodels,GeRMsong}, advanced robotics~\cite{NYUv2,kitti,pst900} and autonomous vehicles~\cite{mfnet,msrs,cmnext}. 
Particularly, the fusion of RGB and Thermal (RGB-T) imagery leverages the complementary characteristics of these sensors, combining the rich texture from RGB cameras with the robustness of Thermal imaging in poor illumination or adverse weather, enabling reliable robotic operations in darkness~\cite{hu2021two}, enhancing autonomous driving under inclement conditions~\cite{liu2019dual}, and facilitating specialized tasks such as cave exploration~\cite{pst900}.
The success of these applications hinges on sophisticated fusion strategies that integrate information from both modalities to achieve performance exceeding that of any single sensor~\cite{CMX, feanet, EAEFNet, CRM, unveiling}.

\begin{figure}
  \centering
  \includegraphics[width=0.5\textwidth]{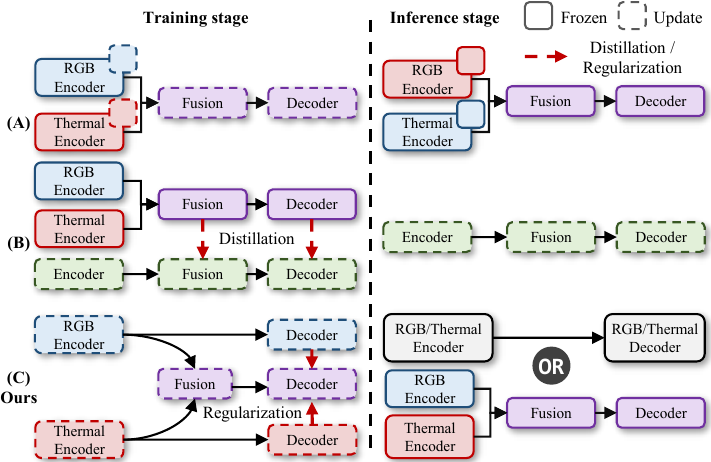}
  \caption{Architecture comparison of training paradigms for RGB–Thermal segmentation under missing modalities. (A) Two-stage knowledge distillation trains a multimodal teacher then distills to per-case students. (B) Modal-adaptation fine-tuning freezes the base and updates lightweight adapters to accommodate a dropped modality. (C) Our unified robustness training jointly optimizes RGB/Thermal encoders, a fusion branch, and multi-branch decoders with bidirectional consistency. At inference, we need to load only the corresponding encoder and decoder parameters.}
  \vspace{-20pt}
\label{teaser}
\end{figure}
However, a fundamental assumption underpins the vast majority of state-of-the-art models, namely that all sensor inputs are consistently available and perfectly aligned.
This assumption is fragile in real-world deployments, where sensors are prone to failure from hardware malfunctions, signal corruption, or environmental interference.


In such cases, how does performance degrade?
As shown in Table~\ref{tab:modal_drop}, we observe a disquieting trend that when one modality is lost, the performance of leading RGB-T segmentation models~\cite{rtfnet, cmnext} drops sharply, often falling below that of a system trained on a single modality from the start.
This reveals that current multimodal systems lack the robustness required for dependable real-world operation in the face of sensor failures.

Traditional approaches~\cite{MS-UDA,CEKD} distill the multimodal knowledge of an RGB-T teacher into a single-modal student (RGB or Thermal).
This paradigm hinders efficient training because it typically requires training separate students for each missing-modality scenario.
To improve the efficiency of teacher-student pipelines, more recent approaches aim to build robustness directly into the model or its training process.
These methods largely fall into two categories.
The first is post hoc, parameter-efficient adaptation~\cite{robust} as shown in Fig.~\ref{teaser}(A). 
In this paradigm, a large pre-trained multimodal model is kept frozen, and lightweight adapters are trained specifically for each missing-modality scenario.
The second is to integrate robustness into the training objective itself~\cite{CRM} as illustrated in Fig.~\ref{teaser}(B), 
where a single model is trained from the start to be resilient to missing inputs (e.g., via self-distillation with complementary masking).

However, despite these advances, significant limitations remain. 
(i) Insufficient feature fusion attention. Adapter-based robustness freezes the multimodal backbone and trains small, scenario-specific adapters.
Because the fusion path gets little to no supervision, adapters hinge on a fixed fused representation, so unimodal branches gain little and remain weak when that representation is suboptimal.
(ii) Insufficient decoupling. Single modality inputs still traverse the fused path, preventing standalone RGB only or T only inference needed for sensor dropouts, safety, and efficient fallback.

\begin{table}[t]
\centering
\setlength{\tabcolsep}{3pt}      
\renewcommand{\arraystretch}{0.98} 
\caption{Comparison on RGB-T segmentation with modality drops, (including signal loss, image corruption, and signal instability).}
\begin{tabular}{l c cc cc}
\toprule
\multirow{2}{*}{\textbf{Methods}} 
  & \multicolumn{1}{c}{\textbf{RGB-T}} 
  & \multicolumn{2}{c}{\textbf{RGB drop}} 
  & \multicolumn{2}{c}{\textbf{Thermal drop}} \\
 & $\text{mIoU}\,\uparrow$ 
 & $\text{mIoU}\,\uparrow$ & \textbf{Diff}\,$\downarrow$ 
 & $\text{mIoU}\,\uparrow$ & \textbf{Diff}\,$\downarrow$ \\
\midrule
RTFNet(ResNet152) & 53.2 & 37.3 & {\color{red}-15.9} & 24.57 & {\color{red}-28.63} \\
EAEFNet(ResNet152) & 58.95 & 35.23 & {\color{red}-23.72} & 41.72 & {\color{red}-17.23}\\
CMXNet(segformer-b4) & 59.77 & 53.55 & {\color{red}-6.22} & 35.46 & {\color{red}-24.31} \\
StitchFusion(segformer-b2) &58.04 &48.78 & {\color{red}-9.26} & 41.42 & {\color{red}16.62}\\
CRM(swin-T)   & 59.1 & 50.98 & {\color{red}-8.12} & 50.22 & {\color{red}-8.88}\\
\midrule
Ours(segformer-b2) & 58.97 &{55.12} & {\color{red}\textbf{-3.85}} &\textbf{53.2} &{\color{red}\textbf{-5.77}} \\
\bottomrule
\end{tabular}
\label{tab:modal_drop}
\vspace{-20pt}
\end{table}

To tackle these limitations, in this paper we propose \method, which combines the \textbf{\underline{F}}usion and \textbf{\underline{D}}ecouple strategy in a three-branch architecture, as illustrated in Fig.~\ref{teaser}(C).
\textbf{Our goal is a representation-level reversible pipeline: forward complementary fusion strengthens the fused stream, while reverse decoupling preserves modality-recoverable components to guide each unimodal branch for standalone fallback.}
The key design includes three modules. Synergistic Feature Fusion (SFF) enriches each single-modality branch with complementary cues through selective cross-modal interaction, and Cross-modal Decouple Regularization (CMDR) explicitly separates modality-specific features from shared ones. Region Decouple Regularization (RDR) ensures that all branches produce semantically consistent outputs.

These components create a unified feedback loop that jointly optimizes the RGB and Thermal encoders, the fusion branch, and their respective decoders.
Specifically, at the inference stage, one simply loads the encoder and decoder corresponding to the available modality.
This design not only preserves the high accuracy of the full-modality system but also markedly improves its robustness with missing or corrupted inputs.

Our contributions are threefold:
\begin{itemize}
\item A novel fusion and decoupling method that combines RGB and Thermal information and \textbf{preserves modality-recoverable components in the fused representation for reverse guidance}.
\item An efficient and inference-capable, parameter-separable, three-branch encoder-decoder network assembled with the proposed feature fusion and decoupling strategy.
\item State-of-the-art performance on three RGB-T semantic segmentation benchmark datasets.
\end{itemize}

The remainder of this paper is organized as follows. Section II reviews related work. Section III details the proposed framework and fusion strategy. Section IV presents both quantitative and qualitative evaluations. Section V concludes the study.
\begin{figure*}
  \centering
  \includegraphics[width=1\textwidth]{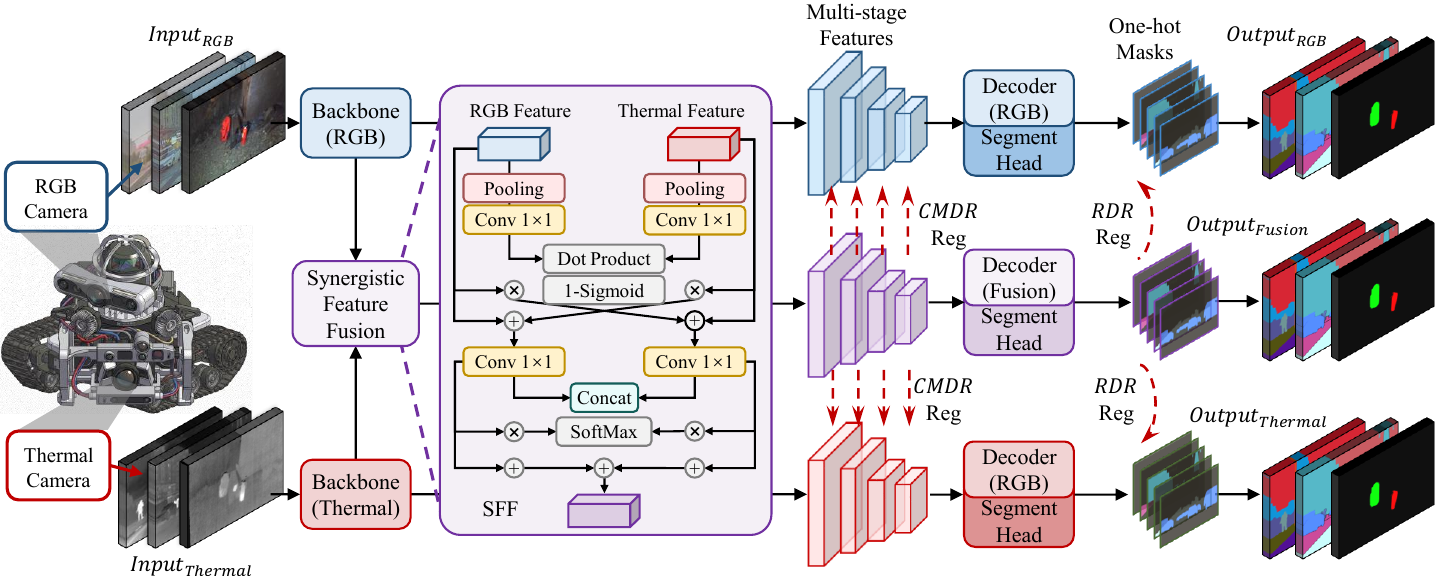}
  \caption{Diagram of encoder–decoder model where SFF executes modality fusion, and CMDR/RDR decouple the fused features to regularize and guide each single-modality branch. (The robotic platform is a conceptual illustration.)}
  \vspace{-15pt}
\end{figure*}

\label{overall}

\section{related work}

\subsection{Feature Fusion in RGB-T Semantic Segmentation}
Leveraging Thermal images to complement RGB images has shown great improvements in semantic segmentation accuracy.
Previous modality fusion methods such as \cite{mfnet,rtfnet,fuseseg} integrated through direct feature addition, verifying the effectiveness of the feature fusion strategy.
Recent work focuses on enhancing finer-grained feature fusion to facilitate more comprehensive cross-modal interaction. Methods such as \cite{abmdrnet,EAEFNet,crossfuse,gmnet,egfnet,mffenet} promote interplay between modalities through more elaborate, dedicated fusion modules.
Other approaches like \cite{HAPNet,CMX,cmnext,fusionsam,stitchfusion} leverage the powerful long-range dependency modeling capabilities of Transformers~\cite{vit} to perform feature interaction or alignment concurrently with feature extraction for different modalities, thereby better promoting modality fusion.
While cross-modal interactions have improved segmentation, reliance on complete inputs makes fusion models brittle, and when a modality is missing, performance collapses.

\subsection{Decoupling for Missing Modalities}
To mitigate the brittleness of feature-fusion methods under modality loss, prior work decouples cross-modal knowledge from a fusion model and transfers it to unimodal subnetworks \cite{MS-UDA,CEKD,2025Distilling}. Two paradigms are prevalent. KD-decouple methods (e.g., CRM~\cite{CRM}, LCM~\cite{lcm}) use the fusion model as a teacher to distill modality-specific students. Adaptation-decouple methods (e.g., Adapted~\cite{robust}) freeze the pretrained fusion model and train lightweight, modality-aware adapters for specific missing-modality scenarios. Collectively, these strategies aim to maintain accuracy with incomplete inputs by strengthening unimodal competence.
In contrast to prior work, our work proposes a fusion–decoupling mechanism to simultaneously strengthen the fusion backbone and achieve inference-time per-modality decoupling.

\section{Method}
\subsection{Framework Overview}
We employ a three-branch encoder-decoder architecture for RGB-Thermal semantic segmentation. 
The RGB and Thermal encoders are connected to Synergistic Feature Fusion (SFF), which facilitates multi-scale information complementary and alignment feature to the Fused stream. Three decoders (RGB, fused, Thermal) are jointly trained based on SegFormer~\cite{segformer}.
Additionally, our network introduces two novel regularizer modules to enhance robustness: 
(1) Cross-modal Decouple Regularization (CMDR), which separates shared and modality-specific signals, and 
(2) Region Decouple Regularization (RDR), which enforces prediction consistency between the fused branch and the RGB or Thermal branches in confident regions.

\subsection{Synergistic Feature Fusion (SFF)}
Let $f_{rgb}, f_{t} \in \mathbb{R}^{C \times H \times W}$ be the intermediate features from the Thermal and RGB encoders, respectively, at a certain scale. To capture channel-wise context, we first generate channel-wise descriptors $\mathbf{T}, \mathbf{R} \in \mathbb{R}^{C \times 1 \times 1}$ for each modality:

\begin{equation}
\begin{aligned}
\mathbf{R} &= F_{MLP}\left(0.5 \cdot (F_{GAP}(f_{rgb}) + F_{GMP}(f_{rgb})) \right)\\
\mathbf{T} &= F_{MLP}\left(0.5 \cdot (F_{GAP}(f_t) + F_{GMP}(f_t)) \right)
\end{aligned}
\end{equation}
Here, $F_{GAP}$ and $F_{GMP}$ denote global average pooling and global max pooling, respectively. $F_{MLP}$ is a two-layer perceptron that maps the aggregated features into a compact representation.

Inspired by EAEFNet~\cite{EAEFNet}, we hypothesize that when the RGB and thermal branches have channel-wise attention vectors $\mathbf{R}, \mathbf{T}\in\mathbb{R}^{C}$ with opposite signs on a channel ($\mathbf{R}\odot\mathbf{T}<0$), the two modalities attend to different semantics and thus provide complementary information. We therefore introduce a dynamic gating mechanism that
selectively amplifies cross-modal flow on channels satisfying $\mathbf{R}\odot\mathbf{T}<0$, producing the
initially enhanced features $f'_{\mathrm{rgb}}$ and $f'_{\mathrm{t}}$.

\begin{equation}
\begin{aligned}
\mathbf f_{\mathrm{rgb}}' &= \mathbf f_{\mathrm{rgb}} + (1-\sigma(\mathbf T \odot \mathbf R))\circledast \mathbf f_t,\\
\mathbf f_t' &= \mathbf f_t + (1-\sigma(\mathbf T \odot \mathbf R))\circledast \mathbf f_{\mathrm{rgb}}.
\end{aligned}
\end{equation}
where $\sigma$ denotes the sigmoid function, $\odot$ the element-wise product, and $\circledast$ the channel-wise multiplication. This operation allows each branch to enhance its features by dynamically borrowing salient information from the other.

Finally, the two refined feature streams are concatenated and passed through $1 \times 1$ convolution layer. Formally, the merged features are obtained by
\begin{equation}
    \begin{aligned}
        f^{''}_{rgb,t} = f{'}_{rgb,t} \otimes SoftMax(Conv_{1\times1}(\overline{f^{'}}_{rgb,t}))
    \end{aligned}
\end{equation}
where $\overline{f^{'}}_{rgb,t}$ denotes the concatenated result of $f_{rgb}'$ and $f_t'$. The outputted features of SFF are obtained by:
\begin{equation}
f_{fuse} = f^{''}_{rgb} + f^{''}_{t}
\end{equation}
This two-stage process, combining channel-wise gating and efficient linear spatial attention, ensures a comprehensive and computationally feasible fusion of information across the two modalities.

\subsection{Cross-Modal Decouple Regularization (CMDR)}
\label{sec:cmdr}

The goal of Cross-Modal Decouple Regularization (CMDR) is to leverage the stronger fused representation—benefiting from the synergistic information of both modalities—to decouple modality-specific signals and transfer them to the uni-modal branches. We then impose a regularization that aligns the distributions of uni-modal features with their decoupled counterparts, thereby guiding and improving each decoder branch. This enhances the robustness and performance of the RGB and Thermal branches when used independently.

Concretely, we invert the modality-alignment component in the SFF to derive channel-wise attention from the fused feature $\mathbf F$ with respect to each modality. Building upon the concept of sign-consistency, we propose a novel approach that leverages the relationship between a uni-modal feature and the fused feature as a selection criterion: when $\mathbf R$ (or $\mathbf T$) and $\mathbf F$ activate with the same sign (i.e., $\mathbf R > 0, \mathbf F > 0$ or $\mathbf R < 0, \mathbf F < 0$), they are considered to be in the same state. Extracting the channels of $\mathbf F$ that share the same state with $\mathbf R$ (or $\mathbf T$) effectively isolates the RGB (or Thermal) component embedded in the fused representation. Formally,
\begin{equation}
\begin{aligned}
\mathbf f^{d}_{\mathrm{rgb}} &= \big(\sigma(\mathbf F \odot \mathbf R)\big)\,\circledast\, \mathbf f_{\mathrm{fuse}},\\
\mathbf f^{d}_{\mathrm{t}}   &= \big(\sigma(\mathbf F \odot \mathbf T)\big)\,\circledast\, \mathbf f_{\mathrm{fuse}},
\end{aligned}
\end{equation}
where $\odot$ denotes element-wise multiplication, $\circledast$ denotes channel-wise rescaling (attention), and $\sigma(\cdot)$ is an element-wise sign-consistency gate (mapping positive entries to~1 and non-positive entries to~0). The tensors $\mathbf f^{d}_{\mathrm{rgb}}$ and $\mathbf f^{d}_{\mathrm{t}}$ are the RGB and Thermal features decoupled from the fused feature map.

Let $f_{\mathrm{rgb}}^{(i)}$, $f_{\mathrm{t}}^{(i)}$, $f_{\mathrm{rgb}}^{(i)}$, and $f_{\mathrm{t}}^{(i)}$ denote the feature maps from the $i$-th stage of the fused, RGB, and Thermal decoders, respectively, where $i \in S$ and $S$ is the set of decoder stages where CMDR is applied. We enforce consistency between the uni-modal features and their decoupled targets using an $\ell_2$ loss. To ensure that gradients flow only from the fused branch to the uni-modal branches (and not vice versa), we detach the decoupled targets from the computation graph via a stop-gradient operator $\mathrm{sg}(\cdot)$. The losses are
\begin{equation}
\begin{aligned}
\mathcal{L}_{\mathrm{CMDR}}^{\mathrm{rgb}} &= \sum_{i \in S} \left\| \mathrm{sg}\!\left(f_{\mathrm{rgb}}^{d(i)}\right) - f_{\mathrm{rgb}}^{(i)} \right\|_{2}^{2},\\
\mathcal{L}_{\mathrm{CMDR}}^{\mathrm{t}}   &= \sum_{i \in S} \left\| \mathrm{sg}\!\left(f_{\mathrm{t}}^{d(i)}\right) - f_{\mathrm{t}}^{(i)} \right\|_{2}^{2},
\end{aligned}
\label{eq:cmdr_rgb}
\end{equation}
and the total loss is
\begin{equation}
\mathcal{L}_{\mathrm{CMDR}} = \mathcal{L}_{\mathrm{CMDR}}^{\mathrm{rgb}} + \mathcal{L}_{\mathrm{CMDR}}^{\mathrm{t}}.
\label{eq:cmdr_total}
\end{equation}
By minimizing this objective, CMDR compels each uni-modal decoder to mimic the high-quality, context-aware features distilled from the fused branch. Because the targets are stop-gradient, this supervision improves the standalone RGB and Thermal branches without introducing conflicting gradients into the fusion pathway.

\subsection{Region Decouple Regularization (RDR)}
To improve segmentation performance near object boundaries and other ambiguous regions, we propose \emph{Region Decouple Regularization} (RDR), which aligns the predictions of uni-modal branches to the fused decoder output in a class-specific manner.

\begin{figure}
  \centering
  \includegraphics[width=0.48\textwidth]{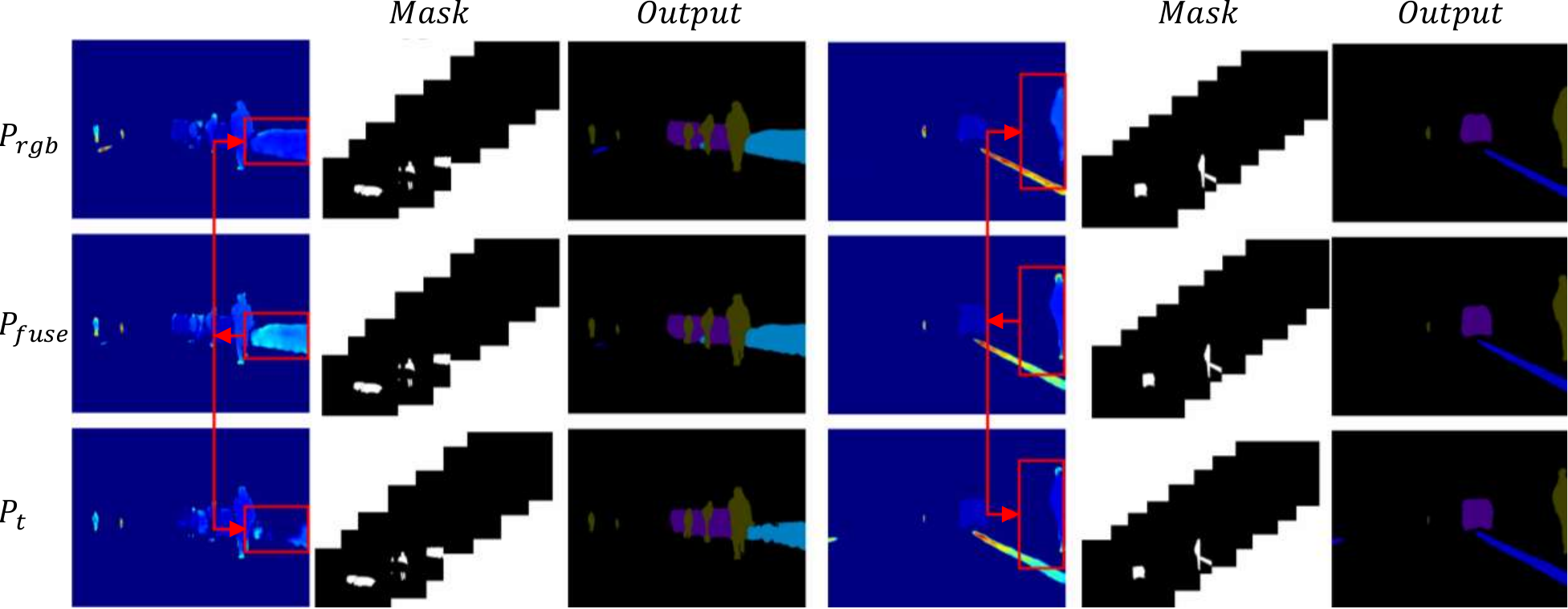}
  \caption{Visualizations of segmentation and feature maps generated by the RGB and Thermal decoders for specific regions.}
\label{cam}
\vspace{-20pt}
\end{figure}

Let $\mathbf{p}_{\text{fuse}}, \mathbf{p}_{\text{rgb}}, \mathbf{p}_{\text{t}} \in \mathbb{R}^{H \times W \times C}$ denote the final probability maps (after softmax) from the fused, RGB, and Thermal decoders, respectively, where $C$ is the number of classes. As shown in Fig.~\ref{cam}, we first convert $\mathbf{p}_{\text{fuse}}$ into one-hot encoded class masks, such that each pixel is assigned to a predicted semantic class:

\begin{equation}
    M = \mathrm{onehot}(\mathbf{p}_{\text{fuse}}, C).
\end{equation}

These class-specific masks are then applied to the fused prediction and used to guide uni-modal outputs via masked, pixel-wise alignment losses:

\begin{equation}
\label{eq:rdr_l1}
\begin{aligned}
\mathcal{L}^{\text{rgb}}_{\mathrm{RDR}} &= \mathcal{L}_{1}\left(\mathrm{sg}(\mathbf{p}_{\text{fuse}} \circledast M),\ \mathbf{p}_{\text{rgb}}\right), \\
\mathcal{L}^{\text{t}}_{\mathrm{RDR}} &= \mathcal{L}_{1}\left(\mathrm{sg}(\mathbf{p}_{\text{fuse}} \circledast M),\ \mathbf{p}_{\text{t}}\right),
\end{aligned}
\end{equation}

where $\mathcal{L}_{1}$ denotes the L1 loss, $\circledast$ indicates element-wise multiplication, and $\mathrm{sg}(\cdot)$ is the stop-gradient operator. This mechanism enforces consistency within confident regions while preventing gradient flow from the fused branch, effectively decoupling modality-specific learning.

The total RDR loss is formulated as:

\begin{equation}
\mathcal{L}_{\mathrm{RDR}} = \lambda_{\mathrm{RDR}} \left( \mathcal{L}^{\text{rgb}}_{\mathrm{RDR}} + \mathcal{L}^{\text{t}}_{\mathrm{RDR}} \right).
\end{equation}

By restricting supervision to class-confident regions and freezing the fused branch via gradient blocking, the fused decoder acts as a fixed guider, while the RGB and Thermal branches learn to align within their respective masked regions.

\subsection{Overall Objective}
The loss function uses the CrossEntropy~\cite{Cross-entropy} for training and the total loss combines CMDR and RDR terms:
\begin{equation}
\mathcal{L}_{ALL} = \lambda \cdot \mathcal{L}_{CMDR} + \lambda^{'} \cdot \mathcal{L}_{RDR} + \lambda'' \cdot \mathcal{L}_{CrossEntropy}
\label{eq:rdr_total}
\end{equation}
By unifying global feature alignment and localized region refinement, our framework robustly adapts to modality degradation, enabling strong uni-modal segmentation performance through cross-modal knowledge transfer.

\section{Experiments}
\label{sec:results}
We performed detailed experiments to evaluate the performance of our proposed method for different tasks and datasets. We also present comparison with existing methods that are robust to missing modalities. 

\subsection{Datasets}
We use three datasets to evaluate the performance of~\method~ on RGB-Thermal segmentation tasks. 

\textbf{MFNet Dataset~\cite{mfnet}} It records 9 semantic categories in urban street scenes, contains 1569 pairs of RGB and Thermal images with a size of $640 \times 480$. Following the split method of RTFNet, we use 784 images for training and 392 images for validation.

\textbf{FMB Dataset~\cite{fmb}} is the first RGB-Thermal dataset with a pixel annotation ratio of greater than 98\%, with 1500 image pairs at the resolution of $800 \times 600$, splitting into 1220/280  for training and testing with 14 categories which often appear in real-world automatic driving tasks.

\textbf{PST900 Dataset~\cite{pst900}} is a dataset for infrared and visible light image, which contains 1,444 pairs of high-quality aligned infrared and visible light images, including 715 daytime image pairs and 729 nighttime image pairs.

\subsection{Implementation Details}
Our~\method~framework adopts a dual-stream architecture built upon SegFormer~\cite{segformer}, where both the teacher and student branches use Mix Transformer (MiT) encoders~\cite{vit} pretrained on ImageNet as backbones. The model is optimized with AdamW, using a weight decay of 0.0001, momentum of 0.9, and initial learning rates of $1 \times 10^{-4}$
for the encoder and $6 \times 10^{-4}$  for the decoder, with cross-entropy loss~\cite{Cross-entropy}. We empirically set the loss balancing weights in Eq.\ref{eq:rdr_total} as $\lambda = 0.5, \lambda^{\prime} = 1.0$, and $\lambda^{\prime\prime} = 1.0$. For a fair comparison, we adopt the same data augmentations as prior methods, including random scaling (0.5–2.0), horizontal flipping, and random-sized cropping.
We train on three datasets and align the resolution and schedule to each. For MFNet we use $480\times640$ RGB--Thermal inputs and train for 300 epochs, and we evaluate using the checkpoint that achieves the best validation performance. For PST we train for 300 epochs at $720\times1280$, while for FMB we also run 300 epochs at $512\times512$.
We primarily use mean Accuracy (mAcc) and mean Intersection-over-Union (mIoU) to quantify segmentation performance. 
To evaluate the robustness of our model, we consider three evaluation settings: (i) both modalities present, (ii) missing RGB input, and (iii) missing Thermal input.

\subsection{Experimental Results}

\begin{table*}[t]
    \centering
    \caption{Performance comparison with existing robust methods for MFNet dataset. 
    RGB and Thermal columns report performance when only RGB and only Thermal are available. 
    RGB-T columns report performance when both modalities are available. 
    Average column reports average performance when one of the two modalities gets missing. 
    `-' indicates that results for those cells are not published. $^\ast$ indicates that available code and pretrained models from the authors were used to generate the results.
   }
    \setlength{\tabcolsep}{6pt}
    \renewcommand{\arraystretch}{1.2}
    \begin{tabular}{llcccccccccc}
        \toprule
        \multirow{2}{*}{Methods} & \multirow{2}{*}{Backbone} & \multirow{2}{*}{Parameters (M)} 
        & \multicolumn{2}{c}{RGB} & \multicolumn{2}{c}{Thermal} & \multicolumn{2}{c}{RGB-T} & \multicolumn{2}{c}{Average} \\ 
        \cmidrule(lr){4-5} \cmidrule(lr){6-7} \cmidrule(lr){8-9} \cmidrule(lr){10-11}
        & & & mAcc & mIoU (\%) & mAcc & mIoU (\%) & mAcc & mIoU (\%) & mAcc & mIoU (\%) \\ 
        \midrule
        FuseNet~\cite{fusenet}     & VGG-16~\cite{vgg}         & 284    & 11.11 & 10.31 & 41.33 & 36.85 & 52.40 & 45.60 & 34.95 & 30.92 \\
        MFNet~\cite{mfnet}         & DCNN~\cite{dcnn}          & 8.4    & 26.62 & 24.78 & 19.65 & 16.64 & 45.10 & 39.70 & 30.46 & 27.04 \\
        MDRNet~\cite{mdrnet}       & ResNet-50~\cite{resnet}   & 64.60  & 57.11 & 45.89 & 41.98 & 30.19 & 74.70 & 56.80 & 57.93 & 44.29 \\
        VPFNet~\cite{vpfnet}       & ResNet-50~\cite{resnet}   & -      & 48.14 & 41.08 & 42.20 & 35.80 & 70.91 & 57.61 & 53.75 & 44.83 \\
        EAEFNet~\cite{EAEFNet}$^\ast$     & ResNet-50~\cite{resnet}   & 200.4  & 37.76 & 35.23 & 49.93 & 41.72 & \textbf{75.07} & 58.95 & 54.25 & 45.30 \\
        RTFNet~\cite{rtfnet}       & ResNet-152~\cite{resnet}  & 254.51 & 44.89 & 37.30 & 26.41 & 24.57 & 55.70 & 53.20 & 42.33 & 38.36 \\
        FEANet~\cite{feanet}       & ResNet-152~\cite{resnet}      & 337.1  & 15.96 &  8.69 & 58.35 & 48.72 & 73.20 & 55.30 & 49.17 & 37.57 \\ 
    
        \midrule
        CRM~\cite{CRM}$^\ast$             & Swin-T~\cite{liu2021swin} & 74.92 & 68.90 & 50.98 & 62.61 & 50.22 & 71.83 & 59.10 & 67.78 & 53.43 \\
        StitchFusion~\cite{stitchfusion}$^\ast$ & Swin-T~\cite{liu2021swin} & 65.27 & 54.83 & 48.78 & 45.91 & 41.42 & 66.05 & 58.04 & 55.60 & 49.41 \\
        HKDNet~\cite{2025Distilling}    & Swin-S~\cite{liu2021swin} & - & \textbf{71.40} & 52.50 & -- & -- & 74.00 & 56.50 & -- & -- \\
        \midrule
        CMNeXt~\cite{cmnext}$^\ast$& MiT-B4~\cite{segformer}   & 116.56 & 60.66 & 53.55 & 38.15 & 35.46 & 68.48 & 59.77 & 55.76 & 49.59 \\
        Adapted~\cite{robust}      & MiT-B4~\cite{segformer}   & -      & 67.18 & 55.22 & 66.70 & 50.89 & -- & -- & -- & -- \\
        \midrule
        Ours (MiT-B2)      & MiT-B2~\cite{segformer}   & 50.0 & 65.19 & 55.12 & \textbf{69.41} & 53.23 & 72.27 & 58.97 & \textbf{68.96} & 55.77 \\
        Ours (MiT-B4)      & MiT-B4~\cite{segformer}   & 124.0 & 65.91 & \textbf{56.06} & 68.73 & \textbf{54.89} & 72.08 & \textbf{60.08} & 68.91 & \textbf{57.01}\\
        \bottomrule
    \end{tabular}
    \label{tab:MFNet-comparison-with-other-models}
\end{table*}
\begin{figure*}[!t]
\centering

\renewcommand{\arraystretch}{1.2}

\setlength{\tabcolsep}{3pt}
\includegraphics[width=1\linewidth]{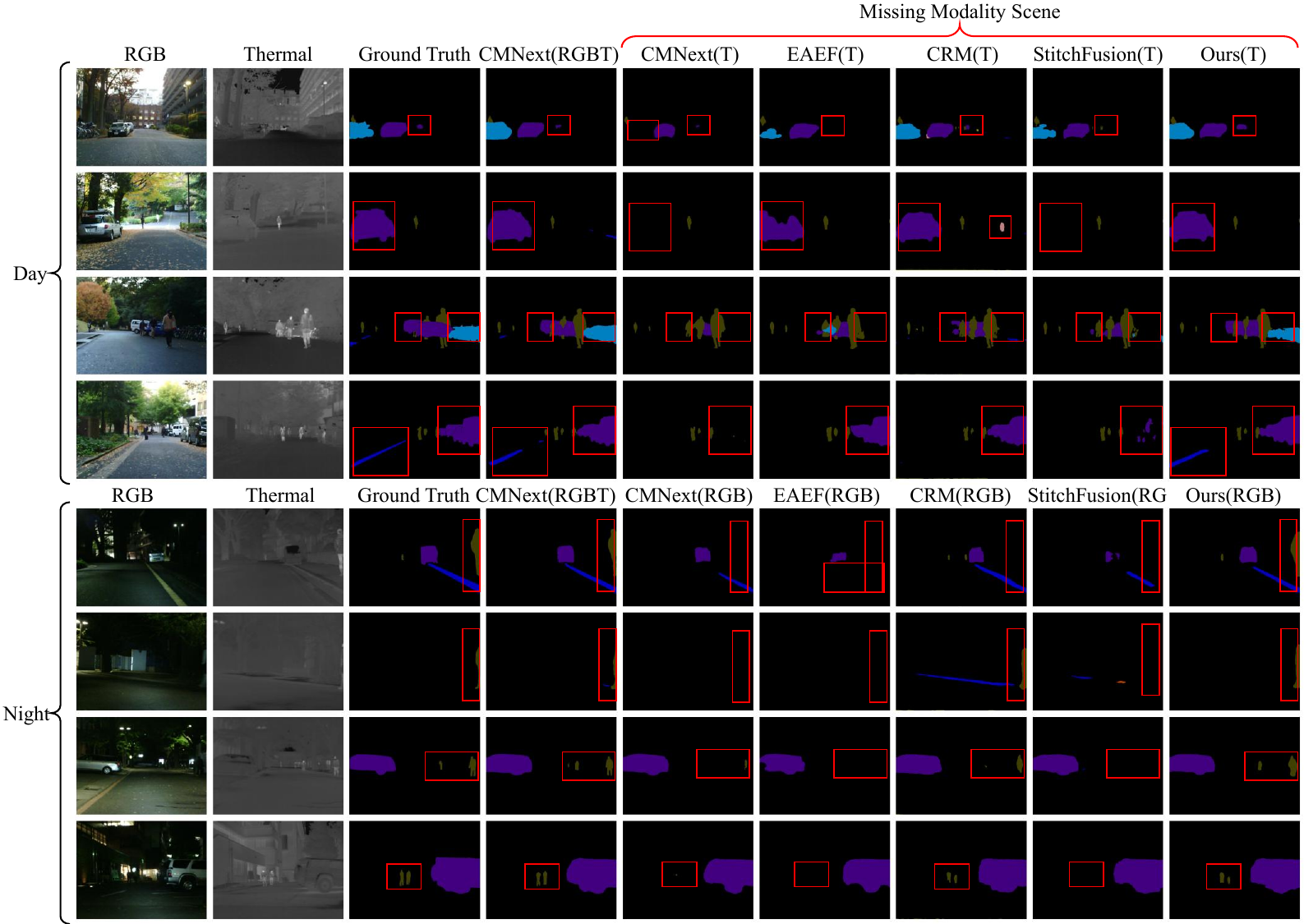}
\caption{The qualitative results on MFNet dataset.}
\label{fig:exp_mf}
\vspace{-10pt}
\end{figure*}


\begin{figure}[!t]
\centering
\includegraphics[width=0.47\textwidth]{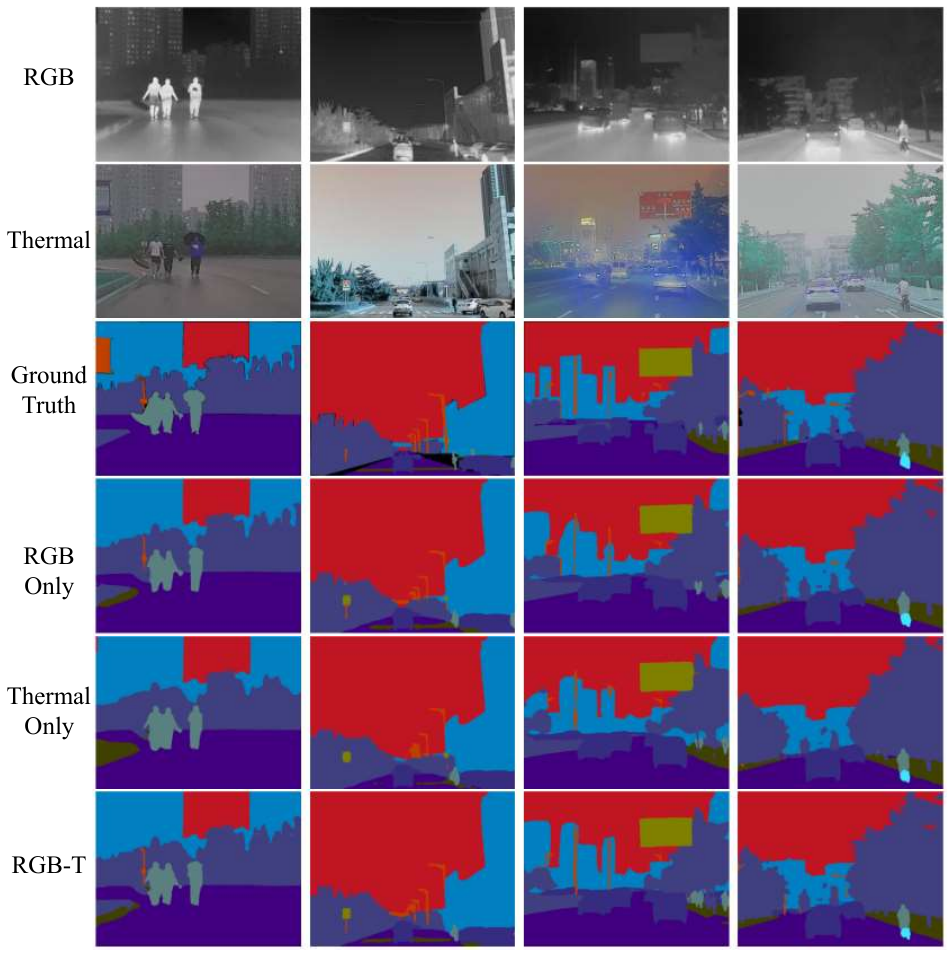}

\caption{The qualitative results on FMB dataset.}
\label{fig_fmb}
\vspace{-10pt}
\end{figure}

\begin{figure}[!t]
\centering
\centering

  \includegraphics[width=0.47\textwidth]{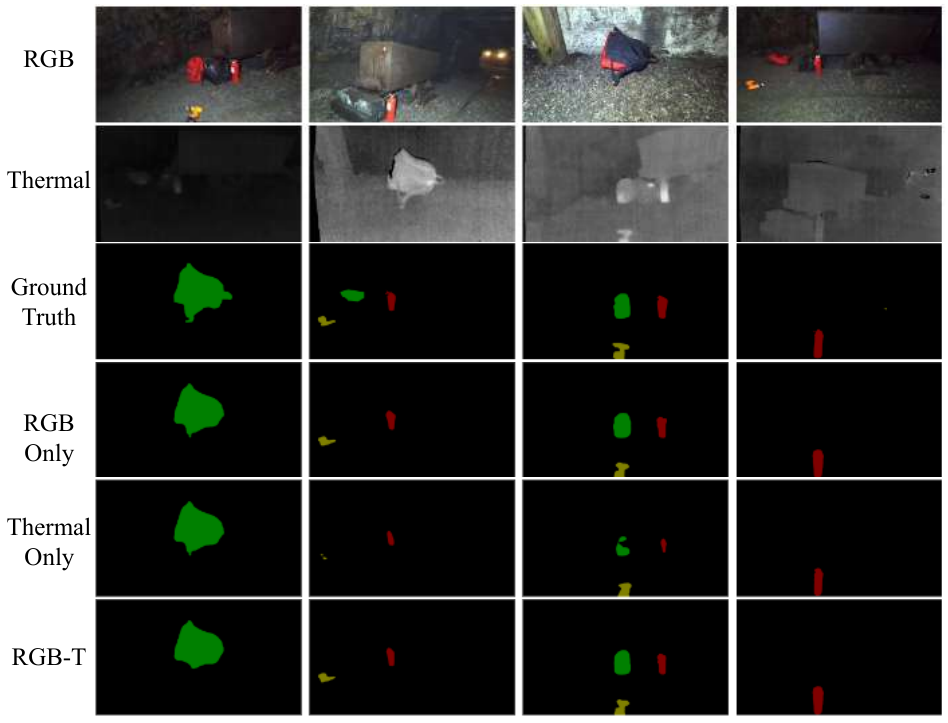}

\caption{The qualitative results on PST900 dataset.}
\label{fig_PST}
\vspace{-15pt}
\end{figure}
\begin{table*}[t]
    \centering
    \caption{Performance comparison with existing robust methods for FMB dataset. 
    RGB and Thermal columns report performance when only RGB and only Thermal are available. RGB-T columns report performance when both modalities are available.
    }
    \label{mf_table}
    \setlength{\tabcolsep}{6pt}
    \renewcommand{\arraystretch}{1.2}
    \begin{tabular}{llcccccccccc}
        \toprule
        \multirow{2}{*}{Methods} & \multirow{2}{*}{Backbone} & \multirow{2}{*}{Parameters (M)} 
        & \multicolumn{2}{c}{RGB} & \multicolumn{2}{c}{Thermal} & \multicolumn{2}{c}{RGB-T} & \multicolumn{2}{c}{Average} \\ 
        \cmidrule(lr){4-5} \cmidrule(lr){6-7} \cmidrule(lr){8-9} \cmidrule(lr){10-11}
        & & & mAcc & mIoU (\%) & mAcc & mIoU (\%) & mAcc & mIoU (\%) & mAcc & mIoU (\%) \\ 
        \midrule
        SegMiF~\cite{fmb} & MiT-B2~\cite{segformer} & - & 73.50 & 50.50 & -- & -- & 74.5 & 54.8 &74.00   &52.65 \\
        StitchFusion~\cite{stitchfusion} & Swin-T~\cite{liu2021swin} & 65.27 & 65.62 & 56.75 & 42.04 & 34.82 & 70.37 & 63.32 & 59.34 & 51.63 \\
        HKDNet~\cite{2025Distilling} & Swin-T~\cite{liu2021swin} & - & 73.5 & 54.0 & - & - & 78.7 & 62.4 &76.1  & 58.2 \\
        \midrule
        \textbf{Ours} (MiT-B2)     & MiT-B2~\cite{segformer}   & 50.0 & 68.57 & 62.00 & 65.11 & 58.98 & 72.07 & 66.17 & 68.58 & 62.38 \\
        \textbf{Ours} (MiT-B4)     & MiT-B4~\cite{segformer}   & 120.0 & \textbf{75.00} & \textbf{67.52} & \textbf{68.72} & \textbf{61.40} & \textbf{76.86} & \textbf{70.73} & \textbf{76.86} & \textbf{66.55} \\
        \bottomrule
    \end{tabular}
    \vspace{-10pt}
    \label{tab:FMB-comparison-with-other-models}
\end{table*}

\begin{table*}[t]
    \centering
    \caption{Performance comparison with existing robust methods for PST900 dataset. 
    RGB and Thermal columns report performance when only RGB and only Thermal are available. RGB-T columns report performance when both modalities are available. 
 }
    
    \setlength{\tabcolsep}{6pt}
    \renewcommand{\arraystretch}{1.2}
    \begin{tabular}{llcccccccccc}
        \toprule
        \multirow{2}{*}{Methods} & \multirow{2}{*}{Backbone} & \multirow{2}{*}{Parameters (M)} 
        & \multicolumn{2}{c}{RGB} & \multicolumn{2}{c}{Thermal} & \multicolumn{2}{c}{RGB-T} & \multicolumn{2}{c}{Average} \\ 
        \cmidrule(lr){4-5} \cmidrule(lr){6-7} \cmidrule(lr){8-9} \cmidrule(lr){10-11}
        & & & mAcc & mIoU (\%) & mAcc & mIoU (\%) & mAcc & mIoU (\%) & mAcc & mIoU (\%) \\ 
        \midrule
        EAEFNet~\cite{EAEFNet}     & ResNet-50~\cite{resnet}   & 200.4  & 88.52 & 83.45 & 21.22 & 20.63 &  91.42 & 85.56 & 67.05 & 63.21 \\
        CRM~\cite{CRM}           & Swin-T~\cite{liu2021swin} & 117.68 & 88.91 & 85.22 & 35.87 & 35.1 & 88.93 & 85.89 & 71.24 & 68.74 \\
        StitchFusion~\cite{stitchfusion}$^\ast$ & Swin-T~\cite{liu2021swin} & - & 87.17 & 82.11 & 20.00 & 19.44 & 90.43 & 84.64 & 65.87 & 62.06 \\
        \midrule
        \textbf{Ours} (MiT-B2)     & MiT-B2~\cite{segformer}   & 120.0 & \textbf{90.28} & 84.3 & \textbf{65.47} & \textbf{62.20} & 87.81 & 84.15 & \textbf{81.19} & \textbf{76.88} \\
        \bottomrule
    \end{tabular}
    \vspace{-10pt}
    \label{tab:PST-comparison-with-other-models}
\end{table*}

As shown in Table \ref{mf_table}, we benchmark~\method~’s performance across multiple datasets under missing-modality conditions.  Our proposed \method demonstrates superior performance and  robustness on the MFNet dataset. In the full-modality (RGB-T) setting, our trained \method (MIT-B4)  achieves a mIoU of 60.08\%, which is competitive with leading methods like CMNeXt and surpasses others. The primary advantage of our approach is highlighted in the missing modalities scenarios.Under missing Thermal inputs, mIoU of our model decreases by 4.02\% on MFNet, achieves 56.06\%, significantly outperforming strong competitors including CMNeXt at 53.55\%, CRM at 50.98\%, and StitchFusion at 48.78\%. In the Thermal-only condition, our model maintains high performance with a mIoU of 54.89\%, only decreases by 5.19\% surpassing most existing methods.

The robustness of our framework is further validated on the FMB and PST900 datasets, as shown in Table \ref{tab:FMB-comparison-with-other-models} and Table \ref{tab:PST-comparison-with-other-models}. On the FMB dataset, our \method achieves  67.52\% in RGB-only setting and 61.40\% in Thermal-only setting, significantly outperforms other methods. A similar trend of significant improvement is observed on the PST900 dataset, where our model consistently achieves the highest scores when missing modalities. 

\subsection{Qualitative Analysis}

For qualitative analysis, Figure \ref{fig:exp_mf} compares our method against several models on the MFNet dataset under modality-incomplete conditions. During daytime, we create a challenging test by evaluating models in a Thermal-only setting, where the absence of rich RGB cues causes significant degradation in most baselines. Our model, however, maintains sharp boundaries and accurate details. For instance, despite weak thermal signatures (row 3), our method fully recovers bicycle silhouettes that competing models fail to detect. Furthermore, our model accurately delineates both the vehicle and the nearly imperceptible curb on the left (row 4). This superiority is even more pronounced in nighttime, RGB-only scenarios (rows 5-8). Here, dark images provide negligible information, causing catastrophic failures in baseline methods. While the majority of models completely fail to detect pedestrians hidden in these severe low-light conditions (rows 5-7), our model consistently and accurately segments them, showcasing its exceptional robustness.

Figure \ref{cam} visualizes the attention maps generated by the three branches of \method in the MFNet dataset. The red arrows and bounding boxes explicitly illustrate the feature transfer process from the multi-modal fusion branch to the respective unimodal branches. A key observation is that even in their respective disadvantageous scenarios such as the RGB branch at night or the Thermal branch in missing texture and color information, the unimodal branches can still focus their attention on the salient object regions identified by the fusion branch. This demonstrates the efficacy of our framework: the capabilities of the fusion branch are effectively transferred to guide the unimodal branches ,significantly enhancing their robustness in missing modalities scenario. 
Furthermore, we present additional qualitative results on the FMB and PST900 datasets in Figure~\ref{fig_fmb} and Figure~\ref{fig_PST}, respectively. Even when restricted to a single modality, our model continues to correctly identify objects on the road and delineate sharp boundaries with only minimal loss of detail.

\subsection{Ablation Study Analysis}
To validate the contribution of each component within \method, we conduct a comprehensive ablation study on the MFNet dataset using the MiT-B2 backbone, as shown in Table \ref{ablstudy}. Starting from a baseline model that utilizes element-wise addition of the two modal features as fusion mechanism without Synergistic Feature Fusion (SFF), it achieves 51.2\% mIoU on the RGB branch and 47.1\% mIoU on the Thermal branch. Replacing the plain fusion with our proposed SFF significantly improves the performance to 53.03\% and 52.68\% mIoU on the RGB and Thermal branches, respectively, demonstrating its superiority in robust feature aggregation over naive fusion strategies. Furthermore, we find that individually incorporating either the CMDR or the RDR into the baseline also yields consistent and significant performance gains. Finally, the full \method framework, which synergistically integrates CGF, CMDR, and RDR, achieves the highest performance, reaching 55.12\% mIoU for the RGB branch and 53.23\% mIoU for the Thermal branch, firmly validating the complementary nature of our fusion-decoupling design.

Table \ref{tab:efficiency} details computational efficiency on an NVIDIA A100. During sensor failure, our architecture seamlessly falls back to a unimodal branch. For MiT-B2, this fallback halves FLOPs (60.9G to 31.5G) and nearly doubles FPS (30.6 to $\sim$58.3), ensuring real-time efficiency for deployable robotics.

\begin{table}[t]
  \centering
  \caption{Results of ablation study on the MFNet dataset.}
  \renewcommand\arraystretch{1}
  \resizebox{\columnwidth}{!}{%
    \begin{tabular}{c|ccc|cc}
    \toprule
    Baseline & SFF & CMDR & RDR & RGB mIoU (\%) & Thermal mIoU (\%) \\
    \hline
    $\surd$ &       &       &       & 51.2  & 47.1 \\
    $\surd$ & $\surd$ &       &       & 53.03 & 52.68 \\
    $\surd$ &       & $\surd$ &       & 52.7  & 49.3 \\
    $\surd$ &       &       & $\surd$ & 53.4  & 50.4 \\
    $\surd$ & $\surd$ & $\surd$ & $\surd$ & \textbf{55.12} & \textbf{53.23} \\
    \bottomrule
    \end{tabular}%
    }
    \vspace{-15pt}
   \label{ablstudy}
\end{table}%

\begin{table}[h]
\centering
\small 
\caption{Efficiency comparison during complete and missing modality inference on NVIDIA A100 (40G).}
\label{tab:efficiency}
\resizebox{\columnwidth}{!}{%
\begin{tabular}{l|c|ccc|c}
\toprule
Backbone & Modality & Params & FLOPs & FPS & mIoU (\%) \\
\midrule
Single B2 & RGB-only & 27.36 M & 31.5 G & 58.3 & 55.12 \\
Single B2 & T-only & 27.36 M & 31.5 G & 58.3 & 53.23 \\
Ours (B2) & RGB+T & 50.0 M & 60.9 G & 30.6 & 58.97 \\
\midrule
Single B4 & RGB-only & 63.2 M & 75.5 G & 27.6 & 56.06 \\
Single B4 & T-only & 63.2 M & 75.5 G & 27.6 & 54.89 \\
Ours (B4) & RGB+T & 124.0 M & 147.0 G & 12.9 & 60.08 \\
\bottomrule
\end{tabular}
}
\vspace{-15pt} 
\end{table}
\section{conclusion}
In this paper, we address the dramatic performance drop in multimodal semantic segmentation under missing modalities by proposing the \method~RGB-T segmentation framework. Unlike traditional robustness methods, \method~unifies efficient feature fusion and modality decoupling in a three-branch architecture, achieving strong segmentation with complete inputs and robust performance when a modality is absent. Specifically, SFF integrates complementary cross-modal cues, while CMDR and RDR strengthen each unimodal branch for standalone inference. This design enables parameter-separable inference and true modality decoupling. Future work will extend this fusion--decoupling framework to additional modalities (e.g., LiDAR or event cameras) and to other perception tasks such as object detection.
{
    \small
    \bibliographystyle{unsrt}
    \bibliography{main}
}

\end{document}